\pdfoutput=1
\documentclass[11pt]{article}
\usepackage[final]{acl}
\usepackage{times}
\usepackage{latexsym}
\usepackage[T1]{fontenc}
\usepackage[utf8]{inputenc}
\usepackage{microtype}
\usepackage{inconsolata}
\usepackage{graphicx}
\usepackage{makecell}
\usepackage{multirow}
\usepackage{booktabs}
\usepackage{array}
\usepackage{nicematrix}
\title{Pre-training data selection for biomedical domain adaptation using journal impact metrics}

\author{Mathieu Laï-king \and Patrick Paroubek \\
  Université Paris-Saclay, CNRS,\\ 
  Laboratoire Interdisciplinaire des Sciences du Numérique,\\ 
  91400, Orsay, France \\
  \texttt{\{mathieu.laiking,patrick.paroubek\}@lisn.upsaclay.fr} 
}

\begin{document}
\maketitle
\begin{abstract}
Domain adaptation is a widely used method in natural language processing (NLP) to improve the performance of a language model within a specific domain. This method is particularly common in the biomedical domain, which sees regular publication of numerous scientific articles. PubMed, a significant corpus of text, is frequently used in the biomedical domain. The primary objective of this study is to explore whether refining a pre-training dataset using specific quality metrics for scientific papers can enhance the performance of the resulting model. To accomplish this, we employ two straightforward journal impact metrics and conduct experiments by continually pre-training BERT on various subsets of the complete PubMed training set, we then evaluate the resulting models on biomedical language understanding tasks from the BLURB benchmark. Our results show that pruning using journal impact metrics is not efficient. But we also show that pre-training using fewer abstracts (but with the same number of training steps) does not necessarily decrease the resulting model's performance.
\end{abstract}

\section{Introduction}
Advances in deep learning for natural language processing (NLP) in recent years have enabled transfer learning to develop \cite{ruderTransferLearningNatural2019}, particularly since the creation of \textit{Transformers} \cite{vaswaniAttentionAllYou2017a}. 

One type of transfer learning aims to start with a pre-training phase where the model learns the general language structure and then a second phase where the model can be fine-tuned for a specific task. In the context of deep learning for NLP, this method avoids re-training a model from scratch for each new task, starting with a model that already has general language knowledge. These pre-trained models generally use a large corpus of text.

A specialized domain, such as finance or the biomedical domain, may contain numerous tasks. In the case of language, a specialized domain has a specific vocabulary containing terms more rarely found in general texts. We can observe this phenomenon when looking at tokens produced by a biomedical tokenizer against a general tokenizer \cite{boukkouriRetrainTrainScratch2022}. Moreover, tasks may require domain-specific knowledge not found in general sources. So, to improve the performance of a model previously trained on a general domain to a specific domain, it is interesting to use a corpus specific to the domain to which we wish to adapt our model. 

Most of the data used for pre-training in the biomedical field are research articles and papers that can be either abstracts, full texts, or a combination of both. This data generally originates from large public databases such as PubMed or PubMedCentral (for full-text articles). However, to our knowledge, no study has examined the selecting subsets of these large databases for pre-training using metrics specific to scientific papers. That leads us to our research questions: Can a language model be adapted to the biomedical domain by efficiently selecting scientific documents in the pre-training data while maintaining or improving its performance ? Is the journal impact factor a good metric to select scientific documents for pre-training ? 

This paper presents our experiments on adapting the pretrained BERT-base model to the biomedical domain. We get the PubMed January 2024 baseline corpus and define different subset configurations using journal impact metrics: h-index \cite{hirschIndexQuantifyIndividual2005} and Scimago Journal Rank or SJR \cite{guerrero-boteFurtherStepForward2012}. We then perform continual pre-training from the BERT-base model \cite{devlinBERTPretrainingDeep2019} and evaluate it on several tasks from the BLURB benchmark \cite{guDomainSpecificLanguageModel2022}.

\section{Related work}
\subsection{Domain-adaptive and domain-specific pre-training for the biomedical domain}
The adaptation of neural models to the biomedical domain has been extensively studied in recent years, focusing on BERT-type models and, more recently, large generative language models. We distinguish two main categories regarding the pre-training data:
\begin{itemize}
    \item \textit{Mixed-domain pre-training}, where the model has seen data from different domains during the pre-training: it can either be a model that has been pre-trained on a general corpus and then trained on in-domain data or a model trained simultaneously on data from multiple domains, such as biomedical and clinical for example \cite{leeBioBERTPretrainedBiomedical2019,beltagySciBERTPretrainedLanguage2019, pengTransferLearningBiomedical2019}.
    \item \textit{Domain-specific pre-training}, where the model only sees data from a single domain during pre-training. The hypotheses are that by using a domain-specific vocabulary, the models learn more accurate representations of specific in-domain terms (that would be divided by the sub-word tokenization with a general corpus) and that it reduces noise introduced by text completely unrelated to the domain \cite{beltagySciBERTPretrainedLanguage2019, boukkouriRetrainTrainScratch2022, lewisPretrainedLanguageModels2020a, guDomainSpecificLanguageModel2022}.
\end{itemize} 

\subsection{Pre-training data quality for large language models}
Several works focus on selecting sequences using quality metrics for pre-training Transformer models in the general domain, particularly with the advent of large language models and the evolution of the size of pre-training datasets for these models \cite{zhouLIMALessMore2023,attenduNLUDataDiets2023,marionWhenLessMore2023,dasDEFTDataEfficient2023}. 

The adaptation of large language models using scientific articles has been largely studied. However, only a few have emphasized the quality of scientific articles used. For the Galactica model \cite{taylorGalacticaLargeLanguage2022}, they only mention applying \emph{"several quality filters, including excluding papers from journals with certain keywords and also excluding papers with a low journal impact factor"}. Most other models that used PubMed or PubMedCentral for pre-training do not mention any specific selection of data at the document level; most focus on preprocessing steps at the content level (bibliography references, authors, figures and tables, etc.) when dealing with full-text articles \cite{luoBioGPTGenerativePretrained2022,wuPMCLLaMAFurtherFinetuning2023, luoBioMedGPTOpenMultimodal2023, chenMEDITRON70BScalingMedical2023}.

\section{Methods} 
\subsection{Methodology}
We use a similar methodology as \citet{marionWhenLessMore2023}, with some small modifications : 

Let $D$ be a large dataset containing documents and $\xi$ a metric assigning a score to a document. We build a subset $P_{c\xi}$ by adding instances that fit our selection criteria $c$ :
\begin{equation}
    P_{c\xi}=\{d_i\in D | c_{0\xi} \leq \xi(d_i)) \leq c_{1\xi} \}
\end{equation}
Where $c_{0\xi}$ and $c_{1\xi}$ are the lower and upper bound for the criteria $c$ and the metric $\xi$. For each metric, we consider two selection criteria: keeping top or middle part of the distribution of the metric\footnote{we do not use the bottom part because in our case, for the SJR metric, more than 25\% of the dataset had the same value : 0, so the percentiles for the bottom part would be 0 but include more than 25\% of the corpus} of $D$ as the data to be kept. This serves as verifying if the model learns better with high quality documents (defined by the metric, for our metrics, higher is better). We keep either 25\% or 50\% of the documents in $D$. So for instance, if we take the 25~\% of the middle part of the distribution for the metric $\xi$, we should compute the 37.5~\% and 62.5~\% percentiles with respect to metric $\xi$, which corresponds to $c_{0\xi}$ and $c_{1\xi}$, and keep the documents between these two percentiles.

Then, we tokenize each document in the subset, and we concatenate them into sequences of length equal to the model's context length. This differs from \citet{marionWhenLessMore2023} as we do the filtering before tokenization (because our metrics are applied on a document, not on a sequence of tokens). These sequences are then used to pre-train a model. Moreover, our metrics do not rely on a reference language model.
The goal is then to pre-train a model on a subset of the whole training set while retaining or improving the model's performance. 

\subsection{Pre-training corpus}\label{sec:methods-corpus}
We use the PubMed Baseline corpus comprising all article abstracts deposited on the PubMed database until January 2024. Using PubMed metadata, we filter out abstracts that are not in English, abstracts whose text is not available, and abstracts whose ISSN journal identifier is not present (we filter this to have enough abstracts with a score as our pruning metrics are based on journal impact). After filtering, the total corpus is comprised of 15.9B tokens.

We did not perform a pre-training experiment using the non-filtered PubMed set because we did not have enough articles with journal identifiers to obtain convenient metric percentiles. Still, we expect this filtering to already impact the overall quality of the corpus.

\subsection{Quality metrics}
The nature of the datasets used for general model training (by which we mean models that are not domain-specific) differs from those used in the biomedical field. They are generally huge datasets comprising texts extracted from the Internet on various sites. In our case, these are research articles from the same database. This presupposes a text quality that is adequate in certain respects (generally correct syntax and formal language, unlike texts found on the Internet). 

We wanted to use metrics specific to scientific articles that have meaning for scientific article readers. So, we decided to use journal impact metrics. We used the metadata available on PubMed. This type of metric can provide insight into the probable impact that a paper can have but does not necessarily ensure scientific quality. However, we believe filtering with impact metrics in a large corpus can help reduce the noise, help the model learn biomedical language, and learn biomedical knowledge more efficiently. We use the h-index \cite{hirschIndexQuantifyIndividual2005} and the SJR \cite{guerrero-boteFurtherStepForward2012} as the data is publicly available on the Scimago website\footnote{\url{https://www.scimagojr.com/journalrank.php}}. For comparison, we also perform a random score assignation on all papers from the dataset; we do not perform multiple random assignations to limit the compute cost.

We computed the percentiles for SJR and h-index and, as there were zero values for the SJR index (for the 12.5\% and 25\% percentiles), we did not perform all the pre-trainings for the \textit{mid} criteria, we only considered the \textit{25~\%} subset. This is also why we did not consider the bottom percentiles. We also did not perform the pre-training on the complete set because of time and resource constraints, but we plan to do it in future work.

\subsection{Pre-processing}
We tokenize the whole dataset and concatenate the text of the different abstracts into sequences of length 512 tokens (maximum sequence length for the model we use: BERT \cite{devlinBERTPretrainingDeep2019}). We keep 5~\% of this set as validation data.

\subsection{Model and pre-training}
We use the original \textit{BERT-base} model \cite{devlinBERTPretrainingDeep2019}, continue pre-training on the defined datasets with masked language modeling, and compare the resulting models. For each pre-training (on each subset), we fix a shared global number of steps so that each model sees the same quantity of tokens: we select the number of steps as the total number needed for one epoch on the entire PubMed corpus. For the runs with the subsets, the model will run multiple epochs until it reaches the total number of steps, with data shuffling between epochs (for example, two epochs for the run where we take the top 50\% of PubMed abstracts with respect to h-index).

We train with a sequence length of 512 and a batch size of 8192\footnote{We perform gradient accumulation and data parallelism to get this batch size.}, which gives us a total of 3598 steps. We use a linear schedule with 10~\% warmup and a peak learning rate of $1e-4$. For the other hyperparameters, we follow the original BERT paper. We train our different models on 2 NVIDIA A100 GPUs.

\begin{table*}[!ht]
    
    \centering
    \begin{NiceTabular}{lcccccccccc}
        \toprule
         & \multirow{2}{*}{base} & \multicolumn{2}{c}{\multirow{2}{*}{random}} & \multicolumn{4}{c}{h-index} & \multicolumn{3}{c}{sjr} \\
         \cmidrule(lr){5-8}\cmidrule(lr){9-11}
         &  & \multicolumn{2}{c}{} & \multicolumn{2}{c}{mid} & \multicolumn{2}{c}{top} & mid & \multicolumn{2}{c}{top} \\
        \cmidrule(lr){2-2}\cmidrule(lr){3-4}\cmidrule(lr){5-6}\cmidrule(lr){7-8}\cmidrule(lr){9-9}\cmidrule(lr){10-11}
         & 0\% & 25\% & 50\% & 25\% & 50\% & 25\% & 50\% & 25\% & 25\% & 50\% \\
        \midrule
        BC5-chem & 87.31 & \underline{90.03} & \textbf{90.24} & 89.40 & 89.93 & 89.51 & 89.52 & 89.72 & 89.61 & 89.89 \\
        BC5-disease & 77.09 & \textbf{81.09} & \underline{80.72} & 81.05 & 80.73 & 80.38 & 80.68 & 81.00 & 80.76 & 80.60 \\
        BC2GM & 75.32 & 79.17 & 79.01 & 79.51 & 79.51 & \underline{79.52} & 79.41 & 78.74 & 79.01 & \textbf{79.87} \\
        JNLPBA & 76.77 & 78.02 & 77.85 & 77.51 & 77.95 & 78.13 & \textbf{78.41} & \underline{78.13} & 78.28 & 77.90 \\
        NCBI-disease & 81.59 & 84.89 & 84.45 & \textbf{85.09} & 84.84 & 84.63 & 84.71 & 84.97 & \underline{84.30} & 84.98 \\\hline
        HoC & 79.22 & 84.41 & 84.74 & 84.72 & 84.56 & \underline{84.83} & 84.71 & 84.54 & \textbf{85.07} & 84.76 \\\hline
        ChemProt & 77.07 & 79.25 & 78.83 & 78.94 & \underline{79.72} & 79.00 & \textbf{79.92} & 78.77 & 79.62 & 78.96 \\
        DDI & \textbf{89.11} & 87.54 & 87.70 & \underline{87.91} & 88.27 & 86.46 & 86.80 & 87.05 & 85.92 & 87.76 \\
        GAD & 76.82 & 78.09 & 78.24 & 78.31 & 77.38 & 77.34 & \textbf{78.39} & \underline{77.42} & 78.35 & 77.00 \\\hline
        BioASQ & 72.19 & \underline{75.93} & 75.63 & 75.63 & 75.24 & 74.84 & \textbf{76.07} & 75.85 & 75.50 & 75.22 \\
        PubMed QA & \textbf{55.24} & 55.20 & 55.20 & 55.16 & 55.12 & 54.78 & 55.16 & \underline{55.20} & 55.22 & 55.20 \\\hline\hline
        Micro avg. & 77.07 & \underline{79.42} & 79.33 & 79.38 & 79.39 & 79.04 & \textbf{79.43} & 79.22 & 79.24 & 79.29 \\
        Macro avg. & 75.89 & 78.56 & 78.55 & \underline{78.59} & 78.53 & 78.25 & \textbf{78.64} & 78.41 & 78.53 & 78.46 \\
        \bottomrule
    \end{NiceTabular}
    \caption{Comparison of the performance of our pretrained models on the different evaluation tasks from the BLURB benchmark \cite{guDomainSpecificLanguageModel2022}. \textit{'base'} model is the BERT$_{BASE}$ model \cite{devlinBERTPretrainingDeep2019} from which we continue the pre-training. For the macro average, we average the datasets from the same task and then average the performance on each task. For each task or average, the \textbf{best performance is in bold} and the \underline{second best performance is underlined}.}
    \label{tab:blurb_results}
\end{table*}

\subsection{Evaluation and fine-tuning}
We evaluate the produced pre-trained models on some of the datasets from the BLURB benchmark \cite{guDomainSpecificLanguageModel2022}. We also re-evaluate the BERT-base model to ensure a consistent evaluation with our fine-tuning scripts. We excluded the PICO and Sentence Similarity tasks (EBM-PICO \cite{nyeCorpusMultiLevelAnnotations2018} and BIOSSES \cite{soganciogluBIOSSESSemanticSentence2017}), for which we had trouble reproducing similar and consistent results across runs to those obtained in the BLURB paper, as they did not share any code to perform the fine-tuning and evaluation. So, we are left with the following evaluation tasks : 
\begin{itemize}
    \item Named entity recognition (NER) : BC5-chem \& BC5-disease \cite{liBioCreativeCDRTask2016}, BC2GM \cite{smithOverviewBioCreativeII2008}, JNLPBA \cite{collierIntroductionBioentityRecognition2004} and NCBI-disease \cite{doganNCBIDiseaseCorpus2014}. We evaluate the models for NER tasks using the \textit{entity-level F1 score}. We model the entities using BIO tags.
    \item Relation extraction : ChemProt\cite{DBLP:journals/biodb/LiSJSWLDMWL16}, DDI \cite{herrero-zazoDDICorpusAnnotated2013}, GAD \cite{bravoExtractionRelationsGenes2015}. We evaluate the models for relation extraction using the \textit{micro F1 score}. We use entity dummyfication with start and end tags and use the [CLS] token to classify relations.
    \item Document classification : HoC \cite{bakerAutomaticSemanticClassification2016}, for which we measure the \textit{micro F1 score}.
    \item Question answering : PubMedQA \cite{jinPubMedQADatasetBiomedical2019a} and BioASQ Task 7b \cite{nentidisResultsSeventhEdition2020}. We evaluate these tasks using \textit{accuracy}.
\end{itemize}

\section{Results and Discussion}
To limit random effects, we perform the fine-tuning multiple times with different random seeds, as described in the BLURB paper: using five seeds for all datasets except for BioASQ and PubMedQA, for which we use ten seeds (because they are smaller in size). We then report the average performance across the different seeds for each dataset in the table~\ref{tab:blurb_results}. 

\subsection{Improvement against non biomedical model}
All models trained on biomedical data perform better than the base model trained only on general-domain data. However, for a fair comparison, we should train it for the same amount of steps on non-biomedical data.

\subsection{Are journal impact metrics important for the model ?}
We obtain the best results in micro and macro averages for the model trained on the top 50\% of the entire set with respect to the h-index of the journal in which abstracts have been published. Overall, the h-index metric performs better than SJR, which may be because the SJR percentile values are very close to each other, so the quality differences are less important.

However, the performance differences are low when we compare to the SJR metric or even when selecting abstracts randomly, regardless of the proportion of abstracts we keep. So, journal impact metrics do not seem important when selecting pre-training data from a corpus of scientific articles. We then should find more appropriate metrics to define the quality of a single abstract or test it on a full-text article corpus (so that the impact of a single document is higher).

\subsection{Is it better to pre-train a model using more abstracts ?}
If we compare the performance difference when training with 25\% of the data against 50\%, we globally have better performances (except for the random selection), but these differences are not significant. So, it would be interesting to perform further pre-training experiments using different subset sizes to investigate which number of documents is optimal for the domain adaptation. 

\section{Conclusion}
This paper presents our early experiments on selecting the pre-training data for the biomedical domain. We show that the journal impact metrics are not better than the random selection at a fixed number of training steps. We also observe that reducing the number of abstracts in the training set does not necessarily decrease the final model performance and show the need to investigate how many documents we need to pre-train a model without losing performance.
 
Further directions include finding better metrics (or combinations of metrics) to assess the quality of a document in the pre-training corpus, investigating metrics at a different level (at the corpus level using various mixtures of biomedical domains), and using a corpus of full-text articles.

\section{Acknowledgments}
This project was provided with computer and storage resources by GENCI at IDRIS thanks to the grant 20XX-AD011014707 on the supercomputer Jean-Zay's A100 partition .
\bibliography{paper}

\end{document}